\documentclass[runningheads]{llncs}

\usepackage{eccv}

\usepackage{eccvabbrv}

\usepackage{graphicx}
\usepackage{booktabs}
\usepackage{tabularx}
\usepackage{tikz}
\usepackage{bm} 

\usepackage{amsmath}
\DeclareMathOperator*{\argmin}{arg\,min} 

\usepackage[accsupp]{axessibility}  

\usepackage{multirow}
\usepackage{colortbl}

\usepackage{hyperref}

\usepackage{orcidlink}

\newcommand*\circled[1]{\tikz[baseline=(char.base)]{\node[shape=circle,draw,inner sep=0.5pt] (char) {#1};}}
\newcommand{\blue}[1]{\textcolor{blue}{#1}}

\begin{document}

\title{Learning to Unlearn for Robust Machine Unlearning} 

\titlerunning{Learning to Unlearn}

\author{
Mark He Huang\inst{1}\orcidlink{0000-0002-9217-4977} \and
Lin Geng Foo\inst{1}\orcidlink{0000-0002-6082-6002} \and
Jun Liu\inst{1,2}\thanks{Corresponding author.}\orcidlink{0000-0002-4365-4165}}

\authorrunning{M.H.~Huang et al.}

\institute{
ISTD Pillar, Singapore University of Technology and Design (SUTD), Singapore\\
\email{\{he\_huang,lingeng\_foo\}@mymail.sutd.edu.sg}
\and
School of Computing and Communications, Lancaster University, UK\\
\email{j.liu81@lancaster.ac.uk}
}

\maketitle


\begin{abstract}
    Machine unlearning (MU) seeks to remove knowledge of specific data samples from trained models without the necessity for complete retraining, a task made challenging by the dual objectives of effective erasure of data and maintaining the overall performance of the model.
    Despite recent advances in this field, balancing between the dual objectives of unlearning remains challenging.
    From a fresh perspective of generalization, we introduce a novel Learning-to-Unlearn (LTU) framework, which adopts a meta-learning approach to optimize the unlearning process to improve forgetting and remembering in a unified manner. 
    LTU includes a meta-optimization scheme that facilitates models to effectively preserve generalizable knowledge with only a small subset of the remaining set, while thoroughly forgetting the specific data samples.     
    We also introduce a Gradient Harmonization strategy to align the optimization trajectories for remembering and forgetting via mitigating gradient conflicts, thus ensuring efficient and effective model updates. 
    Our approach demonstrates improved efficiency and efficacy for MU, offering a promising solution to the challenges of data rights and model reusability.
\end{abstract}

\section{Introduction}
\label{sec:intro}

As artificial intelligence integrates more deeply into our daily lives, machine learning (ML) services are quickly becoming the infrastructure that supports all kinds of user-facing applications \cite{zhou2023unified,zhang2022distilling,blanco2024digital}. 
Yet, the training of these ML models involves gathering vast quantities of user data -- a process that can inadvertently lead to infringing copyright and jeopardizing individuals' privacy. 
Supported by regulations like GDPR~\cite{mantelero2013eu}, users can request data deletion and the service provider is obligated to remove the data from both their database, as well as their trained model weights, which is a non-trivial task.
Although one naive but exact unlearning approach is to retrain the model from scratch on the \textit{remaining set}
(i.e. the original training dataset minus the specific data targeted for removal), 
such full retraining is often impractical due to its significant computational expenses.
Hence, the field of Machine Unlearning (MU) has emerged, 
which aims to make a trained model \textit{forget} certain training data (i.e. the designated \textit{forget set}) at a computational cost less than full retraining \cite{zhang2023review,kurmanji2024towards,nguyen2020variational,xu2024machine,xu2023machine}.

The task of MU has dual intents: (1) \textit{Forgetting}: To forget knowledge learned from the forget set (denoted as $\mathcal{D}_f$) and (2) \textit{Remembering}: To retain knowledge learned from the remaining set (denoted as $\mathcal{D}_r$). However, achieving \textit{forgetting while remembering} is a challenging task because it requires fine adjustments to the model's weights, that selectively erase the influence of the forget set without undermining the accuracy and generalization capabilities derived from the remaining set.
To enable \textit{forgetting} in a trained model, several studies~\cite{golatkar2020eternal,golatkar2020forgetting,foster2023ssd,liu2023muter,mehta2022deep} directly perturb the model weights according to a calculated estimation of the forget set's influence. 
While these approaches can effectively impair the model's ability on $\mathcal{D}_f$, it requires carefully tuning the magnitude of impairment to avoid overly forgetting and causing model performance on $\mathcal{D}_r$ to collapse post-unlearning.
Thus, to make the model \textit{remember} better, a recent line of works\cite{chundawat2023badt,graves2021amnesiac,tarun2023fast,chourasia2022forget,kim2022efficient,fan2023salun} incorporate training or distillation with $\mathcal{D}_r$ into the unlearning process, where direct training signals from $\mathcal{D}_r$ help the model re-learn the knowledge that is potentially removed by the forgetting procedure. 
However, when the original dataset (and $\mathcal{D}_r$) becomes larger in scale, it becomes increasingly difficult and inefficient to obtain feedback from $\mathcal{D}_r$ that helps 
improve \textit{forgetting} and \textit{remembering} at the same time in a cost-effective manner.

Therefore, to tackle the above-mentioned issues and improve MU performance, we approach the problem of \textit{forgetting} while \textit{remembering} from a fresh perspective of generalization.
More precisely, to make the model \textit{remember} better, we aim to obtain feedback from just a small subset of the remaining set $\mathcal{D}_r$ that tells the model what knowledge is \textit{generalizable} and should not be removed during unlearning.
By obtaining feedback on what is generalizable knowledge, we can ensure the forgetting is precisely localized to $\mathcal{D}_f$ and maintain the generalization performance of the model on $\mathcal{D}_r$. 
At the same time, we would also like to improve the \textit{forgetting} of the model, such that the existence of the forget set $\mathcal{D}_f$ is more thoroughly removed from the model.
To achieve this, we leverage signals from membership inference (MI)~\cite{hu2022survey,shokri2017membership,salem2018ml,di2024adversarial} models, which aim to infer whether a data sample exists in the training set.
Specifically, we rely on the insight that various MI models often use different signals to determine if the model has been trained using a specified data sample.
Thus, we can obtain feedback from these MI models on how to facilitate forgetting in a \textit{generalizable} manner, such that the influence of $\mathcal{D}_f$ is removed more thoroughly.

Yet, how to optimize the model during unlearning towards these two generalization goals is a challenging problem.
For instance, it can be difficult to obtain feedback on what is the generalizable knowledge to remember, by using only a small subset of the remaining set $\mathcal{D}_r$.
Moreover, it is also challenging to obtain feedback on how to forget in a comprehensive manner efficiently, by utilizing a few existing MI models.
Hence, to overcome these challenges of generalization, we propose a Learning-to-Unlearn (\textbf{LTU}) framework from a novel perspective of \textit{meta learning}. 
Our LTU framework simultaneously obtains \textit{generalizable} feedback towards \textit{forgetting} and \textit{remembering} in a unified manner, while using only a small subset of $\mathcal{D}_r$.

Meta learning, also known as \textit{``learning to learn''}, involves additional testing during model training for \textit{better generalization capability} \cite{MAML,nichol2018first}. Inspired by this, to enhance the generalization towards remembering and forgetting, we design a meta-optimization scheme in our LTU framework, consisting of three phases: \textbf{meta-tune}, \textbf{meta-test}, and lastly \textbf{meta-update}. 
Intuitively, \textbf{meta-tune} simulates a direct update with the unlearning objective, then \textbf{meta-test} evaluates the temporarily updated model on a separate and disjoint set of samples, which provides feedback on how to optimize in a generalizable manner.
This is followed by \textbf{meta-update}, which consolidates the feedback from the first two phases, and applies the actual unlearning update to the model parameters with improved generalization capabilities.

Besides, the dynamics of MU introduce a challenging scenario, where \textit{remembering} and \textit{forgetting} typically represent conflicting optimization trajectories. 
For instance, optimizing for model utility may inadvertently bias the model towards retaining all knowledge, including that which should be forgone. Conversely, prioritizing forgetting could deter the model from maintaining valuable generalized knowledge. 
To mitigate this issue, we propose a Gradient Harmonization strategy that synchronizes the gradient directions for \textit{remembering} and \textit{forgetting} through gradient projection, which effectively eradicates conflicting elements between the gradients, enabling a cohesive and efficient optimization route that concurrently caters to both unlearning objectives.

\section{Related Work}
\label{sec:related-work}

\subsection{Machine Unlearning}

\textbf{Exact unlearning}
While the naive exact unlearning procedure is to retrain the model from scratch, some previous works~\cite{SISA,koch2023no,dukler2023safe} try to achieve exact unlearning with reduced cost, represented by SISA-based methods~\cite{SISA} making use of the idea of sharding to retrain a sub-model.
While these methods provide a guarantee of data removal comparable to full retraining, they still incur significant computational costs, greatly limiting their applicability.
Differently, we explore the approximate unlearning, which is more practical.
\ \\
\textbf{Approximate unlearning}
Recent works in MU focus on approximate unlearning, which tries to reduce the influence of the target unlearning samples from the trained model while preserving the model's utility.
One line of work~\cite{golatkar2020eternal,golatkar2020forgetting,foster2023ssd,liu2023muter,mehta2022deep} 
operates directly on the model weights without training and is based on estimating the influence of the forget set on the target model, e.g., via Influence Theory~\cite{guo2019certified}, Hessian Matrix~\cite{mehta2022deep,liu2023muter}, and Fisher Information~\cite{golatkar2020eternal,foster2023ssd,peste2021ssse}. 
However, these methods can often cause harm to the model performance post-unlearning.
On the other hand, training-based methods~\cite{fan2023salun,dukler2023safe,chundawat2023badt,chourasia2022forget,tarun2023fast,graves2021amnesiac,wu2020deltagrad,kim2022efficient,koh2023disposable} often maintain better model performance post-unlearning by incorporating training on the full or subset of the \textit{remaining set} within their unlearning procedures.
Yet, these training-based methods can get inefficient when the original dataset gets larger in scale.
Many other studies investigated MU for specific settings \cite{lin2023erm, chen2023boundary, tarun2023fast, kodge2023deep, zhang2024geniu,shen2024label,golatkar2021mixed,zhang2022prompt,ye2022learning,cha2024learning} and applications~\cite{cheng2023gnndelete,heng2024selective,li2024machine, kurmanji2024towards,chen2024fast}, e.g. debias etc.
In contrast, our LTU framework approaches the MU problem via a fresh perspective of generalization to simultaneously perform remembering and forgetting effectively with only a small subset of the remaining set.
\ \\
\textbf{Membership Inference}
Existing studies in MU~\cite{chundawat2023badt,foster2023ssd,fan2023salun,lin2023erm} frequently use membership inference (MI) as a metric for assessing the efficacy of influence removal of the forget set. 
Originally introduced by \cite{shokri2017membership}, MI attacks seek to determine if a particular data sample was part of the training dataset for a ML model by analyzing the model's output patterns. 
Subsequent works further improve MI, e.g., by reducing the requirements for the attack \cite{salem2018ml} and enhancing performance by using distilled loss trajectories \cite{liu2022membership}. 
Notably, \cite{chen2021machine} highlighted that traditional unlearning methods could inadvertently leave traces of the data meant to be forgotten, making them vulnerable to sophisticated MI attacks. Unlike previous works that utilize MI merely for evaluation, our work leverages differentiable MI models to actively guide the unlearning process. Through a meta-optimization framework, we utilize gradient feedback from various MI models to significantly improve the thoroughness of our unlearning method.

\subsection{Meta Learning}

The paradigm of meta-learning \cite{MAML,rajeswaran2019meta,snell2017prototypical,guo2020learning,antoniou2019learning}, which is also known as ``learning to learn'', emerged to mainly tackle the few-shot learning problem.
Specifically, early methods \cite{MAML,nichol2018first} improve the few-shot performance by learning an optimal initialization of network parameters such that the test-time adaptation with a few samples becomes more rapid and effective. 
Recently, the paradigm of meta-learning has extended into other areas \cite{huang2021metasets,xu2023experts,MCRES_2023_CVPR,foo2022era,foo2023system,peng2024harnessing,peng2024joint}, enhancing model generalization via the guidance of second-order gradients without necessitating test-time updates. 
Unlike existing works, we are the first to leverage meta-learning-based techniques to enable effective and efficient machine unlearning.

\section{Method}
\label{sec:method}

\begin{figure}[t] 
   \centering
   \includegraphics[width=\linewidth]{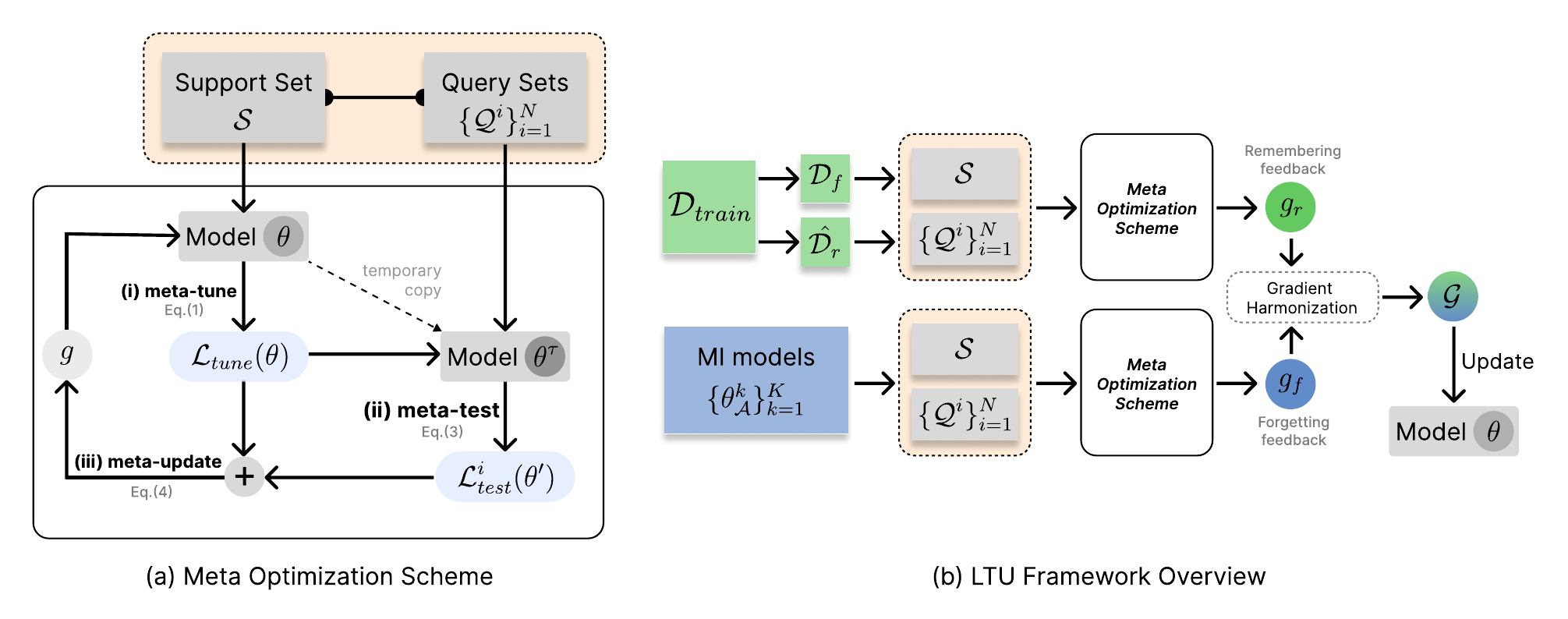}
   \caption{
   (a) Illustration of our meta optimization scheme, which is performed in three phases: \textbf{meta-tune}, \textbf{meta-test} and \textbf{meta-update}.
   Before performing meta optimization, we first construct the support set $\mathcal{S}$ and query sets $\{ \mathcal{Q}^i \}_{i=1}^N$ (highlighted in orange), which are disjoint from each other.
   Then, \textbf{meta-tune} is performed on the support set $\mathcal{S}$ to obtain a temporarily updated model $\theta^\tau$, followed by \textbf{meta-test} evaluations of $\theta^\tau$ using the query sets $\{ \mathcal{Q}^i \}_{i=1}^N$, which provides feedback on the generalization of the ``\textit{remembering}'' or ``\textit{forgetting}''. 
   The feedback from these two phases are combined into \textbf{meta-update}, which computes a gradient to update the model $\theta$ in a generalizable manner.
   (b) Overview of our entire LTU pipeline, which includes one part to improve ``\textit{remembering}'' (top half) and one part to improve ``\textit{forgetting}'' (lower half). Specifically, to improve ``\textit{remembering}'' with only a small subset $\hat{\mathcal{D}_r}$ of the remaining set, we use $\hat{\mathcal{D}_r}$ to construct $N$ query sets.
   In contrast, to improve ``\textit{forgetting}'', we sample from $K$ membership inference models $\{ \theta_{\mathcal{A}}^k \}_{k=1}^K$ to construct our support and query sets.
   Lastly, a Gradient Harmonization strategy is proposed to reduce the conflicts between the gradients from both parts, which further improves unlearning.
   }   
   \label{fig:main}
\end{figure}

Existing MU methods often face challenges balancing the trade-off between \textit{forgetting} and \textit{remembering}. On the one hand, overly \textit{forgetting} will cause the model utility to decline over time, on the other hand, methods that leverage the use of the full remaining set ($\mathcal{D}_r$) for \textit{remembering} are costly and sometimes infeasible.

To tackle these issues and improve MU performance, we approach the \textit{remembering} and \textit{forgetting} problem via the fresh perspective of \textit{generalization} and propose our Learning-to-Unlearn (LTU) framework, which is illustrated in \cref{fig:main}.
Our LTU framework includes a meta optimization scheme (introduced in \cref{sec:meta-optimization-scheme}) that provides feedback regarding generalization capabilities during optimization. 
Then, in order to make the model \textit{remember} better with only a small subset of the remaining data $\mathcal{D}_r$, we apply our meta optimization scheme to provide feedback on what knowledge is generalizable, and avoid removing them during unlearning.
At the same time, in order to \textit{forget} better, 
we apply our meta optimization scheme to obtain feedback from various MI models on how to forget in a more generalizable manner, 
i.e., removing $\mathcal{D}_f$ more thoroughly.
Note that, to achieve these distinct aims, we need to construct appropriate support sets $\mathcal{S}$ and query sets $\mathcal{Q}$ for \textbf{meta-tune} and \textbf{meta-test} respectively (e.g., by intentionally creating gaps between their tasks), which we present in \cref{sec:support-and-query}. 
Therefore, this enables our LTU framework to simultaneously tackle ``\textit{forgetting}'' and ``\textit{remembering}'' in a unified manner.

Additionally, since remembering and forgetting typically represent conflicting optimization trajectories, there could be conflicting elements between the gradients for both objectives, which could reduce the efficacy of the combined framework. Thus, to mitigate this challenge, we propose a Gradient Harmonization strategy in \cref{sec:harmonization} which adopts a gradient projection technique to harmonize the gradients, enabling a cohesive optimization process that concurrently improves both unlearning objectives, thus improving MU performance.

\subsection{Meta Optimization Scheme}
\label{sec:meta-optimization-scheme}

In order to enhance the generalization capabilities in terms of ``\textit{remembering}'' and ``\textit{forgetting}'', we design a meta optimization scheme that consists of three phases: \textbf{meta-tune}, \textbf{meta-test}, and \textbf{meta-update}. 
Specifically, to obtain ``\textit{remembering}'' feedback, we first construct a support set by sampling from $\mathcal{D}_f$ and multiple query sets by sampling from $\mathcal{D}_r$, such that there is a \textit{large gap between the tasks of} handling the support set v.s. handling the query sets (i.e., forgetting task v.s. remembering task).
Then, during meta optimization, we first temporarily tune the target unlearning model by one step with the support set (\textbf{meta-tune}), and then conduct temporary testing on various query sets (\textbf{meta-test}). 
Crucially, since we construct the support and query sets to be different tasks, performing well for \textbf{meta-test} is challenging and requires remembering of only the generalizable knowledge that can enable good performance on the query sets (sampled from $\mathcal{D}_r$) while learning to forget $\mathcal{D}_f$.
Thus, the gradients from \textbf{meta-test} provides feedback which forces the model to \textit{optimize towards remembering generalizable knowledge} during unlearning (\textbf{meta-update}).
Below, going along with the intuition for ``\textit{remembering}'' presented above, we formally introduce our meta optimization scheme based on the intuition for obtaining the ``\textit{remembering}'' feedback, but we remark that the presented formulations and intuitions hold the same for the ``\textit{forgetting}'' feedback as well.

--- \textbf{Meta-tune:} In this step, we fine-tune the target model (i.e. the trained model targeted for unlearning) by one step with the support set $\mathcal{S}$. 
Denoting the model weights as $\theta$, we can compute the \textbf{meta-tune} loss $\mathcal{L}_{tune}$ as:
\begin{equation}\label{eq:tune_loss}
\mathcal{L}_{tune}(\theta) = \mathcal{L}(\theta, \mathcal{S})
\end{equation}
where $\mathcal{L}$ is the cross-entropy loss function, which guides the model to forget $\mathcal{D}_f$ in this case. 
Then, based on the calculated loss $\mathcal{L}_{tune}$, we make a temporary copy of the original model $\theta$ and update its parameters as follows:
\begin{equation}\label{eq:tune_step}
\theta^{\tau} = \theta - \alpha\nabla_\theta \mathcal{L}_{tune}(\theta)
\end{equation}
where $\alpha$ denotes the learning rate during \textbf{meta-tune}. 
Intuitively, this meta-tune step \textit{simulates a direct update} to improve forgetting of the support set $\mathcal{S}$ (i.e., samples from $\mathcal{D}_f$).
Yet, this straightforward approach harms the model performance significantly because the knowledge required for $\mathcal{D}_f$ and $\mathcal{D}_r$ are often entangled, so the model will tend to also forget the generalizable knowledge required for $\mathcal{D}_r$.
Therefore, we do not make an actual update to the target model $\theta$ yet, instead, we update the parameters of a temporary copy of the model $\theta$ to obtain $\theta^\tau$, which will serve as an intermediary for the subsequent \textbf{meta-test}.

--- \textbf{Meta-test:}
In this step, we test the performance of $\theta^{\tau}$ on the constructed query sets $\{\mathcal{Q}^i\}_{i=1}^{N}$, which are from a subset of the remaining set $\hat{\mathcal{D}_r}$ and have a large task gap to the query sets.
Concretely, for each $i$-th query set ($\mathcal{Q}^i$), the loss for \textbf{meta-test} step ($\mathcal{L}_{test}(\theta^{\tau})$) is given by:
\begin{equation}\label{eq:test_loss}
\mathcal{L}_{test}^i(\theta^{\tau}) = \mathcal{L}(\theta^{\tau}, \mathcal{Q}^i).
\end{equation}
Intuitively, this step evaluates the temporarily updated model's ($\theta^{\tau}$'s) performance on a different task, i.e., \textit{remembering}, as required by the query sets.
Importantly, performing well for \textbf{meta-test} after applying updates to forget $\mathcal{D}_f$ (in \textbf{meta-tune}) is challenging, which requires the model to \textit{remember only the generalizable knowledge} that can maintain good performance on $\mathcal{D}_r$.
Thus, the meta-test loss provides feedback which forces the model to localize the forgetting to $\mathcal{D}_f$, optimizing it towards \textit{remembering generalizable knowledge}.
Moreover, as we employ different sampling functions (described in \cref{eq:sample_func1,eq:sample_func2,eq:sample_func3}), we construct the $N$ query sets with diverse distributions, and aggregate the meta-test losses from all of them (i.e., resulting in $\sum_{i=1}^N \mathcal{L}_{test}^i(\theta^{\tau})$).
By using query sets with diverse distributions, this facilitates the \textbf{meta-test} phase to provide more generalizable feedback which can handle data from various distributions.
These test losses ($\sum_{i=1}^N \mathcal{L}_{test}^i(\theta^{\tau})$) are used as feedback to drive the final unlearning objective in the subsequent \textbf{meta-update}.

--- \textbf{Meta-update:}
At \textbf{meta-update} step, we make the \textit{actual} update to the model parameters $\theta$.
Specifically, we combine the feedback obtained in the first two steps to formulate the final optimization objective as follows:
\begin{equation}\label{eq:optimization}
   \begin{aligned}
        & \min_{\theta}\;\mathcal{L}_{tune}(\theta) + \sum_{i=1}^N \mathcal{L}_{test}^i(\theta^{\tau}) \\
      = & \min_{\theta}\;\mathcal{L}_{tune}(\theta) + \sum_{i=1}^N \mathcal{L}_{test}^i \big (\theta - \alpha\nabla_\theta \mathcal{L}_{tune}(\theta) \big)
   \end{aligned}
\end{equation}
where the first term evaluates a straightforward unlearning objective and the second term provides feedback on \textit{how much generalizable knowledge was retained} during \textbf{meta-tune}, based on the performance of $\theta^{\tau}$.
Based on this objective, we update the model parameters ($\theta$) as:
\begin{equation}\label{eq:update}
   \begin{aligned}
      \theta \leftarrow \theta - \beta \nabla_{\theta} \Big(\mathcal{L}_{tune}(\theta) + \sum_{i=1}^N \mathcal{L}_{test}^i \big (\theta - \alpha\nabla_\theta \mathcal{L}_{tune}(\theta) \big) \Big)
   \end{aligned}
\end{equation}
where $\beta$ is the learning rate for \textbf{meta-update}. 
With this update, we can optimize towards making the model forget $\mathcal{D}_f$, while providing it feedback on what is generalizable knowledge which it should not forget in order to maintain its generalization performance on $\mathcal{D}_r$.

\subsection{Support and Query Formation}
\label{sec:support-and-query}

In \cref{sec:meta-optimization-scheme} above, we introduce our meta optimization framework, which can facilitate the generalization of \textit{remembering} and \textit{forgetting} in a unified manner. 
To achieve each of these two aims with meta optimization, we carefully construct their support set $\mathcal{S}$ and query sets $\{\mathcal{Q}^i\}_{i=1}^{N}$ correspondingly (as shown in \cref{fig:main}). 
Specifically, to improve \textit{remembering} while using only a small subset ($\hat{\mathcal{D}}_r$) of the remaining set, we use $\hat{\mathcal{D}}_r$ to construct multiple query sets which are carefully designed to have distributional gaps between them, which facilitates the retaining of generalizable knowledge via meta optimization.
Then, to \textit{forget} $\mathcal{D}_f$ more tracelessly, we construct the support and query sets by sampling from a small set containing various MI models, thus obtaining feedback signals during meta optimization that encourages more generalized forgetting.
Below, we introduce the formation of support and query sets in detail.

\subsubsection{Support and query for remembering feedback}

As we wish to obtain generalized ``\textit{remembering}'' feedback that tells the model what knowledge to not remove at the unlearning time, we construct the support and query sets based on the forget set ($\mathcal{D}_f$) and a small subset of the remaining set ($\mathcal{D}_r$) which we denote as $\hat{\mathcal{D}_r}$. 
By using a small subset, we can ensure the computational cost is low, and unlearning can be done flexibly without the assumption that $\mathcal{D}_r$ is fully accessible at the time of a data-deletion request.
To quantify the relative size of $\hat{\mathcal{D}_r}$ to $\mathcal{D}_r$, we use a constant $\rho \in (0, 1)$ such that $|\hat{\mathcal{D}_r}| = \rho \cdot | \mathcal{D}_r|$.
We first construct the support set $\mathcal{S}$ by drawing samples from the forget set and assigning them with labels drawn randomly from the label space ($Y$), where these random labels represent the forgetting of these samples: 
\begin{equation}\label{eq:sample_df}
   \begin{aligned}
      \mathcal{S} = \{(x_i, y_{rand}) \mid (x_i, y_i) \in \mathcal{D}_f, y_{rand} \in Y\}
   \end{aligned}
\end{equation}
We also sample data from the remaining set ($\hat{\mathcal{D}_r}$) to form the query sets as:
\begin{equation}\label{eq:sample_dr}
   \begin{aligned}
      \{\mathcal{Q}^i\}_{i=1}^{N} = \{(x_i, y_i) \mid (x_i, y_i) \in \mathbf{Z}^i(\hat{\mathcal{D}_r}, k)\}
   \end{aligned}
\end{equation}
where $\mathbf{Z}^i$ produces a subset of a given dataset using a sampling function $z^i$ and drawing $k$ number of samples.
Note that, since $\mathcal{D}_f$ and $\mathcal{D}_r$ are disjoint and have a task gap (i.e., forgetting v.s. remembering), our support set $\mathcal{S}$ and query sets $\{\mathcal{Q}^i\}_{i=1}^{N}$ are disjoint and have a task gap as well, which facilitates our meta optimization framework.
Then, to obtain more effective and generalizable feedback, we want to ensure each query set captures a different data distribution, which can represent different sub-tasks, that helps better evaluation of the generalization capability during \textbf{meta-test}. 
Thus, we employ three different sampling functions while constructing $\{\mathcal{Q}^i\}_{i=1}^{N}$ to aid remembering. 

Our first sampling function is based on the intuition that visually or semantically closer samples of $\hat{\mathcal{D}_r}$ to $\mathcal{D}_f$ tend to share more common knowledge, and are thus more prone to be forgotten along with $\mathcal{D}_f$. Hence, these closer samples can provide stronger signals on what is the generalizable knowledge to retain while forgetting $\mathcal{D}_f$.
Specifically, the first sampling function $z^1 (\cdot)$ draws samples from $\hat{\mathcal{D}_r}$ that are closely related to the samples from $\mathcal{D}_f$ in the feature space:
\begin{equation}\label{eq:sample_func1}
      z^1(\hat{\mathcal{D}_r}, (x_i, y_i)) := \argmin_{(x_j, y_j)} \| \mathcal{F}(x_i) - \mathcal{F}(x_j) \|_2, \text{  s.t. } (x_i, y_i) \in \mathcal{D}_f, (x_j, y_j) \in \hat{\mathcal{D}_r}
\end{equation}
where $\mathcal{F}$ represents a lightweight feature extractor. Note that, since sampling is only conducted within $\hat{\mathcal{D}_r}$ instead of the whole of $\mathcal{D}_r$, the computational cost is still low and does not need to scale with $\mathcal{D}_r$.
Using the same intuition, the second sampling function $z^2 (\cdot)$ will also draw similar samples from $\hat{\mathcal{D}_r}$, but instead measure similarity through comparing samples in the label space ($Y$):
\begin{equation}\label{eq:sample_func2}
      z^2(\hat{\mathcal{D}_r}, (x_i, y_i)) := \argmin_{(x_j, y_j)} \| \mathcal{I}(y_i) - \mathcal{I}(y_j) \|_2, \text{  s.t. } (x_i, y_i) \in \mathcal{D}_f, (x_j, y_j) \in \hat{\mathcal{D}_r}
\end{equation}
where $\mathcal{I}$ represents a generic vectorization function that can be used to determine the distance between labels.
Lastly, we also wish to include samples that have varying distances to the samples in $\mathcal{D}_f$, such that the generalizability to a wider range of samples can be tested during \textbf{meta-test}.
Therefore, the third sampling function $z^3 (\cdot)$ draws samples randomly:
\begin{equation}\label{eq:sample_func3}
    \begin{aligned}
        z^3(\hat{\mathcal{D}_r}) := {random\_pick}(\hat{\mathcal{D}_r})
    \end{aligned}
\end{equation}
By crafting the query sets to be diverse while having task gaps from $\mathcal{D}_f$, we can make the \textbf{meta-test} feedback more generalizable, which facilitates the retaining of generalizable knowledge.
Overall, by developing appropriate support and query sets, we can improve the model's \textit{remembering} by deriving \textit{generalizable} feedback during the process of meta-optimization.

\subsubsection{Support and query for forgetting feedback}

Additionally, we wish to also obtain a generalized ``\textit{forgetting}'' feedback that tells the model how to remove the knowledge of the forget set more thoroughly.
To achieve this, we rely on membership inference (MI) models  \cite{shokri2017membership} that aim to infer whether a data sample was used in training the ML model, which has often been used as a metric to measure the efficacy of forgetting.
Crucially, our insight here is that various MI models often rely on different signals to determine if the model has been trained using a specific data sample, and thus we can leverage them to create task and distributional gaps between the support and query sets to ensure the generalization of the forgetting.

Specifically, for a model $\theta$ that is trained on some dataset $\mathcal{D}$, a data sample $x_i \in \mathcal{D}$ is regarded as a member, conversely, $x_j \notin \mathcal{D}$ is a non-member. Thus, we can abstract the MI attacker's inference model to a binary classifier $\theta_{\mathcal{A}}$ with class probabilities as input and binary labels as output.
Since different $\theta_{\mathcal{A}}$ captures different knowledge about the data membership characteristics, we therefore 
employ a set of $K$ pre-trained fully-differentiable MI models ($\{ \theta_{\mathcal{A}}^k \}_{k=1}^K$) with different model architectures and training processes.
Concretely, at the unlearning time, to facilitate the MI models, we construct an audit set as $\mathcal{D}_{Au} = \{ (x_i, y) \mid x_i \in \mathcal{D}_f \}$.
we pass $\mathcal{D}_{Au}$ through $\theta$ during the forward pass to obtain the prediction logits, which are then passed to the MI model ($\theta_{\mathcal{A}}$) as inputs. 
Then, by employing a binary cross entropy (BCE) loss, we can obtain \textit{forgetting gradient} ($g_{o}$) w.r.t. $\theta$ that can be used as forgetting guidance during unlearning, as follows:
\begin{equation}\label{eq:mi_gradient}
      g_{o} = \nabla_{\theta}\mathcal{L}_{BCE}(\theta_{\mathcal{A}}, \theta(\mathcal{D}_{Au}))
\end{equation}
However, such forgetting guidance is only specific to a single MI model.
To obtain generalized forgetting guidance, we now set up the support-query sets consisting of the different MI models ($\{ \theta_{\mathcal{A}}^k \}_{k=1}^K$). 
At every iteration, we randomly select one MI model to become the support set ($\mathcal{S} \leftarrow \theta_{\mathcal{A}}^i$) and a different MI model to be the query set ($\mathcal{Q} \leftarrow \theta_{\mathcal{A}}^j \mid j \neq i$), thus creating task and distributional gaps between the support and query sets.
In this way, via our meta optimization scheme (\cref{sec:meta-optimization-scheme}), we can obtain an optimized gradient $g_f$ as a \textit{generalizable} feedback that guides the ``\textit{forgetting}'' to be more thorough.

\subsection{Gradient Harmonization}
\label{sec:harmonization}

Above, we discuss the meta-optimization scheme (\cref{sec:meta-optimization-scheme}) and how we construct the support and query sets (\cref{sec:support-and-query}) to improve the unlearning process by obtaining generalizable feedback w.r.t. ``\textit{remembering}'' and ``\textit{forgetting}''.
Yet, it can still be challenging to simultaneously learn to retain and remove, which typically represent conflicting optimization trajectories. 
For example, optimizing to remember generalizable knowledge can potentially push the model to retain all knowledge in general and interfere with forgetting of $\mathcal{D}_f$. 
Conversely, optimizing to forget $\mathcal{D}_f$ thoroughly could encourage the model to forget all knowledge, deterring the model from maintaining valuable generalized knowledge.

Therefore, inspired by previous works \cite{yu2020gradient,pan2023gradmdm} in multi-task learning,
we introduce a \textit{Gradient Harmonization} strategy that harmonizes the gradient directions for ``\textit{remembering}'' and ``\textit{forgetting}'' through gradient projection. This Gradient Harmonization method effectively reduces gradient conflicts, enabling a cohesive and efficient optimization route that concurrently caters to both unlearning objectives. 

\begin{figure}[t] 
   \centering
   \includegraphics[width=0.85\linewidth]{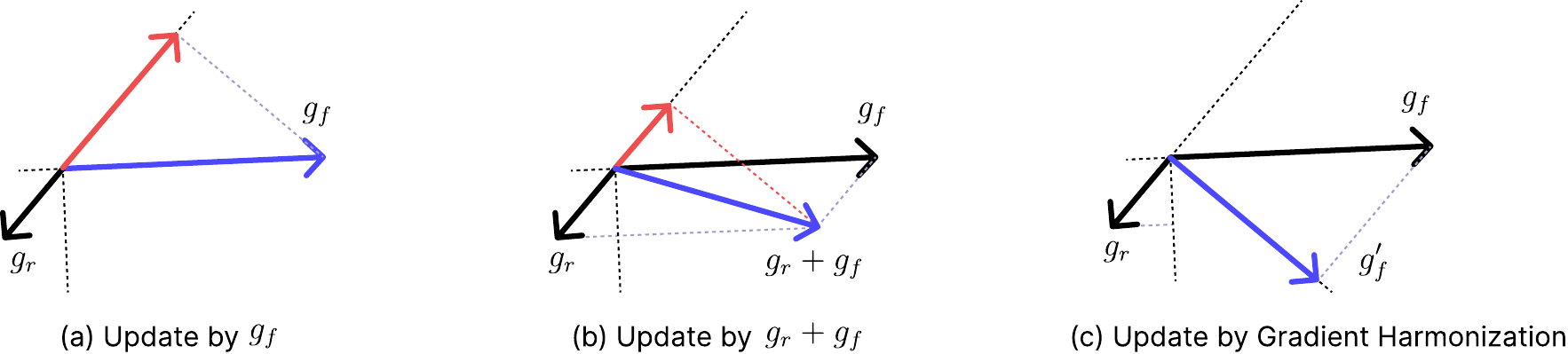}
   \caption{
   Illustration of our Gradient Harmonization strategy.
   (a) Updating with $g_f$ directly might be harmful, as $g_f$ might be in a largely conflicting direction from $g_r$ (highlighted in 
\textcolor{red}{red}), which can result in undoing much of the effects of $g_r$.
   (b) Updating jointly with $g_r + g_f$ can still result in the joint gradient canceling the effect of one of the gradient updates.
   (c) By using our Gradient Harmonization strategy,  we resolve the gradient conflicts, where $g_f^\prime$ is in an orthogonal direction to $g_r$ which reduces the negative impact on the remembering objective.   
   }
   \label{fig:harmonize}
\end{figure}

More precisely, 
we apply our \textit{meta-optimization scheme} (\cref{sec:meta-optimization-scheme}) twice and obtain two optimized gradient updates per iteration (as shown in \cref{fig:main} (b)).
However, these gradients -- one ($g_r$) which aims for better ``\textit{remembering}'' and one ($g_f$) which aims for better ``\textit{forgetting}'' -- can often be conflicting, which can cause issues when optimized simultaneously.
For example, a simple and naive approach would be to directly add the gradients for $g_r$ and $g_f$, to obtain a final gradient for updating the target model.
Yet, this approach can be suboptimal, since the conflicting gradients can often cancel each other out and reduce the overall efficacy (see \cref{fig:harmonize}).
To address this, we propose to harmonize the gradients for our two main objectives via gradient projection, as detailed next.

Initially, for the two gradient vectors $g_r, g_f$ computed by the meta optimization scheme, we compute the cosine similarity between them as $\cos(g_r, g_f) = \frac{g_r \cdot g_f}{\|g_r\| \|g_f\|}$.
If the cosine similarity is less than zero, this means the two gradient vectors have conflicting components, and thus we harmonize them.
More precisely, to harmonize $g_f$ with respect to $g_r$, we first compute the component of $g_f$ that conflicts with $g_r$ as $\frac{g_f \cdot g_r }{\|g_r\| } \frac{g_r}{\|g_r\|}$.
Then, by removing the conflicting component, we can project the gradient vector $g_f$ onto the subspace orthogonal to $g_r$ to obtain $g_f^{\prime}$. The orthogonality implies that updating with this projected gradient $g_f^{\prime}$ will lead to less negative impact for the other objective.
On the other hand, if the initially computed cosine similarity is non-negative, this means that the two gradient vectors are not conflicting, thus we do not need to harmonize them.
These steps are formalized in \cref{eq:proj}:
\begin{equation} \label{eq:proj}
   \begin{split}
      g_f^{\prime} & = g_f - \frac{g_f \cdot g_r}{\|g_r\| \|g_r\|} g_r \text{ , if  } \cos(g_r, g_f) < 0 \\
                   & = g_f \text{ , otherwise; } \\
   \end{split}
\end{equation}
Finally, with the aim of performing thorough ``\textit{forgetting}'' without removing the knowledge required for ``\textit{remembering}'', 
we linearly combine the harmonized gradient vector $g_{f}^{\prime}$ with $g_{r}$ to get the final harmonized gradient as: $\mathcal{G} = g_{r} + g_{f}^{\prime}$.
Overall, $\mathcal{G}$ is the harmonized result of our meta gradients $g_{r}$ and $g_{f}$ with the conflicting component removed, such that we do not interfere with the ``\textit{remembering}'' of generalizable knowledge while optimizing towards thorough ``\textit{forgetting}''. 
In this way, our unlearning feedback is more coherent and effective.

\section{Experiments}
\label{sec:experiments}

\subsection{Experimental Setups}

\textbf{Unlearning Settings.} 
We focus our experiments on \textit{random sample unlearning} as it presents the most challenging and realistic scenario. Note that our method is also seamlessly applicable to sub-class and whole-class unlearning.
\ \\
\textbf{Datasets and Models.} Following \cite{golatkar2020eternal,chundawat2023badt}, we primarily evaluate our method on image classification using CIFAR dataset \cite{CIFAR100}, which contain 50k training images and 10k test images.
Following \cite{lin2023erm,fan2023salun}, we also evaluate on Tiny-ImageNet~\cite{deng2009imagenet}, which contains 100k training images and 10k val images.
Following \cite{chundawat2023badt,foster2023ssd}, we use ResNet-18 \cite{resnet} and Vision Transformer (ViT) \cite{dosovitskiy2020image} for the model architecture. 
\ \\
\textbf{Baselines.} We compare LTU with the following baselines:
(1) \textbf{Retrain}: Full retraining of the model from scratch with the remaining set ($\mathcal{D}_r$). This is exact unlearning and is usually regarded as the ``Gold Model'' (most desired outcome) of MU. Following most existing works \cite{fan2023salun}, we will report and compare the gap between the method result and the gold model.
(2) \textbf{FT}: Fine-tuning the original model with $\mathcal{D}_r$ only.
(3) \textbf{RandL} (Random Label): First update the dataset by re-assigning the labels randomly for the forget set ($\mathcal{D}_f$), then, fine-tune the original model with the updated dataset. 
Besides, we also compare with 6 top-performing representative methods \cite{thudi2022unrolling,graves2021amnesiac,golatkar2020eternal,foster2023ssd,chundawat2023badt,fan2023salun} 
(more details about them in supp).
\ \\
\textbf{Evaluation Measures.} Following common practice in previous works \cite{fan2023salun,foster2023ssd,chundawat2023badt}, we use the following 5 metrics to evaluate the performance of unlearning: 
(1) \textbf{UA} (Unlearning Accuracy): Error on the forget set ($1 - $ accuracy on $\mathcal{D}_f$). 
(2) \textbf{RA} (Remaining Accuracy): Accuracy on the remaining set ($\mathcal{D}_{r}$). 
(3) \textbf{TA} (Testing Accuracy): Accuracy on the original dataset's test set. 
(4) \textbf{MI} (Membership Inference): Accuracy of leveraging MI attack for samples in $\mathcal{D}_f$. We use the same MI model as \cite{fan2023salun} for evaluation and we have purposely excluded this model in our forgetting guidance (detailed in \cref{sec:support-and-query}).
(5) \textbf{UT} (Unlearn Time): Execution time of the unlearning procedures measured in minutes.

Overall, \textbf{UA} \& \textbf{MI} measure how well the model forgets; \textbf{RA} \& \textbf{TA} measure how well the model retains; \textbf{UT} measures computation cost. Note that we compute a delta ($\Delta$) against the Gold Model (\textbf{Retrain} baseline) performance for metrics (1-4). A lower delta indicates better unlearning performance. Good and balanced performance across all five metrics is essential to demonstrate a unlearning method's efficacy and practicality.

\subsection{Experimental Results}

\begin{table}[h]
    \caption{Comparison with 9 baseline methods on CIFAR10 with ResNet-18 (top), and Tiny-ImageNet with Vision Transformer (bottom). The reported results show the mean and standard deviation (shown as $\mu_{\pm std}$) of 10 repeated experiments with different random seeds. The absolute difference ($\Delta$) between the method result and the gold model (Retrain) is presented in parentheses, the lower the gap the better the unlearning performance. All values are in percent (\%) except UT, which is in mins. $\rho$ represents the percentage of $\mathcal{D}_r$ directly used at unlearning time.
    } 
    \label{tab:main}
    \centering
    \resizebox{\textwidth}{!}{
        \begin{tabular}{c|ccccc|ccccc}
        \midrule
        \multirow{2}{*}{\textbf{Methods}} & \multicolumn{5}{c|}{\textbf{ Random Sample Unlearning (10\%) }}  & \multicolumn{5}{c}{\textbf{ Random Sample Unlearning (50\%) }} \\
        & \multicolumn{1}{c|}{UA (\blue{$\Delta \downarrow$})} & \multicolumn{1}{c|}{RA (\blue{$\Delta \downarrow$})} & \multicolumn{1}{c|}{TA (\blue{$\Delta \downarrow$})} & \multicolumn{1}{c|}{MI (\blue{$\Delta \downarrow$})} & UT $\downarrow$ & \multicolumn{1}{c|}{UA (\blue{$\Delta \downarrow$})} & \multicolumn{1}{c|}{RA (\blue{$\Delta \downarrow$})} & \multicolumn{1}{c|}{TA (\blue{$\Delta \downarrow$})} & \multicolumn{1}{c|}{MI (\blue{$\Delta \downarrow$})} & UT $\downarrow$ \\
        \midrule
        Retrain ($\rho=1$) & $5.24_{\pm 0.69}$ (\blue{0.00}) & $100.00_{\pm 0.00}$ (\blue{0.00}) & $94.26_{\pm 0.02}$ (\blue{0.00}) & $12.88_{\pm 0.09}$ (\blue{0.00}) & $43.29$ & $7.91_{\pm 0.11}$ (\blue{0.00}) & $100.00_{\pm 0.00}$ (\blue{0.00}) & $91.72_{\pm 0.31}$ (\blue{0.00}) & $19.29_{\pm 0.06}$ (\blue{0.00}) & $23.90$ \\
        \midrule
        FT ($\rho=1$) & $0.63_{\pm 0.55}$ (\blue{4.61}) & $99.88_{\pm 0.08}$ (\blue{0.12}) & $94.06_{\pm 0.27}$ (\blue{0.20}) & $2.70_{\pm 0.01}$ (\blue{10.18}) & $2.37$ & $0.44_{\pm 0.37}$ (\blue{7.47}) & $99.96_{\pm 0.03}$ (\blue{0.04}) & $94.23_{\pm 0.03}$ (\blue{2.51}) & $2.15_{\pm 0.01}$ (\blue{17.14}) & $6.79$ \\
        RandL ($\rho=1$) & $7.61_{\pm 0.31}$ (\blue{2.37}) & $99.67_{\pm 0.14}$ (\blue{0.33}) & $92.83_{\pm 0.38}$ (\blue{1.43}) & $37.36_{\pm 0.06}$ (\blue{24.48}) & $2.64$ & $4.80_{\pm 0.84}$ (\blue{3.11}) & $99.55_{\pm 0.19}$ (\blue{0.45}) & $91.31_{\pm 0.27}$ (\blue{0.41}) & $41.95_{\pm 0.05}$ (\blue{22.66}) & $6.65$ \\
        GA~\cite{thudi2022unrolling} & $0.69_{\pm 0.54}$ (\blue{4.55}) & $99.50_{\pm 0.38}$ (\blue{0.50}) & $94.01_{\pm 0.47}$ (\blue{0.25}) & $1.70_{\pm 0.01}$ (\blue{11.18}) & $0.13$ & $0.40_{\pm 0.33}$ (\blue{7.51}) & $99.61_{\pm 0.32}$ (\blue{0.39}) & $94.34_{\pm 0.01}$ (\blue{2.62}) & $1.22_{\pm 0.00}$ (\blue{18.07}) & $0.66$ \\
        Amnesiac~\cite{graves2021amnesiac} & $7.58_{\pm 0.77}$ (\blue{2.34}) & $98.87_{\pm 0.67}$ (\blue{1.13}) & $91.98_{\pm 0.34}$ (\blue{2.28}) & $3.76_{\pm 0.97}$ (\blue{9.12}) & $8.92$ & $14.99_{\pm 0.41}$ (\blue{7.08}) & $75.21_{\pm 0.75}$ (\blue{24.79}) & $74.45_{\pm 0.95}$ (\blue{17.27}) & $1.94_{\pm 0.56}$ (\blue{17.35}) & $6.24$ \\
        Fisher~\cite{golatkar2020eternal} & $86.95_{\pm 0.45}$ (\blue{81.71}) & $20.42_{\pm 0.01}$ (\blue{79.58}) & $19.01_{\pm 0.12}$ (\blue{75.25}) & $7.13_{\pm 0.03}$ (\blue{5.75}) & $20.15$ & $77.09_{\pm 0.24}$ (\blue{69.18}) & $22.68_{\pm 0.41}$ (\blue{77.32}) & $22.93_{\pm 0.08}$ (\blue{68.79}) & $8.51_{\pm 0.96}$ (\blue{10.78}) & $11.20$ \\
        SSD~\cite{foster2023ssd} & $9.91_{\pm 0.00}$ (\blue{4.67}) & $87.87_{\pm 0.00}$ (\blue{12.13}) & $82.19_{\pm 0.00}$ (\blue{12.07}) & $5.56_{\pm 0.00}$ (\blue{7.32}) & $2.78$ & $18.18_{\pm 0.00}$ (\blue{10.27}) & $89.47_{\pm 0.00}$ (\blue{10.53}) & $80.25_{\pm 0.00}$ (\blue{11.47}) & $11.40_{\pm 0.00}$ (\blue{7.89}) & $5.79$ \\
        BadT~\cite{chundawat2023badt} ($\rho=1$) & $17.66_{\pm 0.87}$ (\blue{12.42}) & $98.44_{\pm 0.89}$ (\blue{1.56}) & $83.59_{\pm 0.70}$ (\blue{10.67}) & $27.70_{\pm 0.03}$ (\blue{14.82}) & $2.32$ & $27.58_{\pm 0.29}$ (\blue{19.67}) & $80.71_{\pm 0.43}$ (\blue{19.29}) & $79.25_{\pm 0.22}$ (\blue{12.47}) & $39.66_{\pm 0.67}$ (\blue{20.37}) & $6.43$ \\
        SalUn~\cite{fan2023salun} ($\rho=1$) & $4.09_{\pm 0.04}$ (\blue{1.15}) & $99.41_{\pm 0.11}$ (\blue{0.59}) & $93.19_{\pm 0.07}$ (\blue{1.07}) & $12.58_{\pm 0.01}$ (\blue{0.30}) & $2.66$ & $7.54_{\pm 0.22}$ (\blue{0.37}) & $98.06_{\pm 0.17}$ (\blue{1.94}) & $89.93_{\pm 0.20}$ (\blue{1.79}) & $23.08_{\pm 0.01}$ (\blue{3.79}) & $2.68$ \\
        \midrule
        \textbf{LTU ($\bm{\rho=0.3}$)} & $4.37_{\pm 0.55}$ (\blue{0.87}) & $99.83_{\pm 0.09}$ (\blue{0.17}) & $93.95_{\pm 0.39}$ (\blue{0.31}) & $12.97_{\pm 0.02}$ (\blue{0.09}) & $2.63$ & $7.88_{\pm 0.35}$ (\blue{0.03}) & $99.12_{\pm 0.23}$ (\blue{0.88}) & $90.77_{\pm 0.12}$ (\blue{0.95}) & $17.39_{\pm 0.11}$ (\blue{1.90}) & $2.23$ \\
        \bottomrule
        \end{tabular}
    }
    \centering
    \resizebox{\textwidth}{!}{
        \begin{tabular}{c|ccccc|ccccc}
        \midrule
        \multirow{2}{*}{\textbf{Methods}} & \multicolumn{5}{c|}{\textbf{ Random Sample Unlearning (10\%) }}  & \multicolumn{5}{c}{\textbf{ Random Sample Unlearning (50\%) }} \\
        & \multicolumn{1}{c|}{UA (\blue{$\Delta \downarrow$})} & \multicolumn{1}{c|}{RA (\blue{$\Delta \downarrow$})} & \multicolumn{1}{c|}{TA (\blue{$\Delta \downarrow$})} & \multicolumn{1}{c|}{MI (\blue{$\Delta \downarrow$})} & UT $\downarrow$ & \multicolumn{1}{c|}{UA (\blue{$\Delta \downarrow$})} & \multicolumn{1}{c|}{RA (\blue{$\Delta \downarrow$})} & \multicolumn{1}{c|}{TA (\blue{$\Delta \downarrow$})} & \multicolumn{1}{c|}{MI (\blue{$\Delta \downarrow$})} & UT $\downarrow$ \\
        \midrule
        Retrain ($\rho=1$) & $34.24_{\pm 0.00}$ (\blue{0.00}) & $99.98_{\pm 0.00}$ (\blue{0.00}) & $68.23_{\pm 0.00}$ (\blue{0.00}) & $16.86_{\pm 0.00}$ (\blue{0.00}) & $102.98$ & $37.80_{\pm 0.00}$ (\blue{0.00}) & $99.80_{\pm 0.00}$ (\blue{0.00}) & $55.74_{\pm 0.00}$ (\blue{0.00}) & $14.79_{\pm 0.00}$ (\blue{0.00}) & $66.49$ \\
        \midrule
        FT ($\rho=1$) & $32.70_{\pm 1.47}$ (\blue{1.54}) & $87.97_{\pm 1.16}$ (\blue{12.01}) & $63.97_{\pm 0.74}$ (\blue{4.26}) & $19.98_{\pm 1.28}$ (\blue{3.12}) & $5.90$ & $34.44_{\pm 0.99}$ (\blue{3.36}) & $86.20_{\pm 1.02}$ (\blue{13.60}) & $48.30_{\pm 0.95}$ (\blue{7.44}) & $10.91_{\pm 1.36}$ (\blue{3.88}) & $9.54$ \\
        RandL ($\rho=1$) & $62.39_{\pm 1.34}$ (\blue{28.15}) & $42.73_{\pm 0.73}$ (\blue{57.25}) & $33.64_{\pm 0.07}$ (\blue{34.59}) & $21.48_{\pm 0.41}$ (\blue{4.63}) & $5.25$ & $63.42_{\pm 0.00}$ (\blue{25.62}) & $32.18_{\pm 1.11}$ (\blue{67.62}) & $28.72_{\pm 1.23}$ (\blue{27.01}) & $9.31_{\pm 0.84}$ (\blue{5.48}) & $10.53$ \\
        GA~\cite{thudi2022unrolling} & $81.50_{\pm 0.27}$ (\blue{47.26}) & $16.94_{\pm 0.53}$ (\blue{83.04}) & $14.38_{\pm 1.44}$ (\blue{53.85}) & $7.01_{\pm 0.22}$ (\blue{9.85}) & $2.03$ & $88.04_{\pm 0.80}$ (\blue{50.24}) & $13.17_{\pm 0.77}$ (\blue{86.63}) & $10.13_{\pm 0.75}$ (\blue{45.61}) & $22.53_{\pm 0.90}$ (\blue{7.75}) & $8.35$ \\
        Amnesiac~\cite{graves2021amnesiac} & $80.59_{\pm 2.59}$ (\blue{46.35}) & $14.65_{\pm 0.78}$ (\blue{85.33}) & $13.79_{\pm 2.01}$ (\blue{54.44}) & $4.69_{\pm 0.58}$ (\blue{12.17}) & $12.76$ & $82.85_{\pm 3.34}$ (\blue{45.05}) & $20.54_{\pm 0.05}$ (\blue{79.26}) & $9.45_{\pm 3.09}$ (\blue{46.29}) & $20.06_{\pm 1.32}$ (\blue{5.28}) & $14.29$ \\
        Fisher~\cite{golatkar2020eternal} & $93.36_{\pm 0.83}$ (\blue{59.12}) & $20.49_{\pm 2.43}$ (\blue{79.49}) & $19.64_{\pm 0.22}$ (\blue{48.59}) & $9.29_{\pm 2.32}$ (\blue{7.56}) & $49.02$ & $90.05_{\pm 3.42}$ (\blue{52.25}) & $11.57_{\pm 0.23}$ (\blue{88.23}) & $7.36_{\pm 2.98}$ (\blue{48.37}) & $12.13_{\pm 1.81}$ (\blue{2.66}) & $35.79$ \\
        SSD~\cite{foster2023ssd} & $85.23_{\pm 0.00}$ (\blue{50.99}) & $23.81_{\pm 0.00}$ (\blue{76.17}) & $23.88_{\pm 0.00}$ (\blue{44.35}) & $19.20_{\pm 0.00}$ (\blue{2.35}) & $6.20$ & $83.48_{\pm 0.00}$ (\blue{45.68}) & $24.99_{\pm 0.00}$ (\blue{74.81}) & $18.27_{\pm 0.00}$ (\blue{37.47}) & $20.12_{\pm 0.00}$ (\blue{5.33}) & $7.41$ \\
        BadT~\cite{chundawat2023badt} ($\rho=1$) & $52.35_{\pm 0.23}$ (\blue{18.11}) & $59.75_{\pm 0.80}$ (\blue{40.23}) & $51.15_{\pm 0.33}$ (\blue{17.08}) & $22.24_{\pm 0.53}$ (\blue{5.39}) & $4.63$ & $56.88_{\pm 1.36}$ (\blue{19.07}) & $70.00_{\pm 1.40}$ (\blue{29.80}) & $42.12_{\pm 1.08}$ (\blue{13.62}) & $22.02_{\pm 1.05}$ (\blue{7.24}) & $6.82$ \\
        SalUn~\cite{fan2023salun} ($\rho=1$) & $32.32_{\pm 0.21}$ (\blue{1.92}) & $94.07_{\pm 0.85}$ (\blue{5.91}) & $63.55_{\pm 0.76}$ (\blue{4.68}) & $19.23_{\pm 0.89}$ (\blue{2.37}) & $4.48$ & $34.39_{\pm 1.26}$ (\blue{3.41}) & $93.89_{\pm 0.48}$ (\blue{5.91}) & $50.80_{\pm 1.27}$ (\blue{4.94}) & $17.99_{\pm 0.90}$ (\blue{3.20}) & $5.47$ \\
        \midrule
        \textbf{LTU ($\bm{\rho=0.3}$)} & $33.44_{\pm 0.51}$ (\blue{0.80}) & $96.47_{\pm 0.54}$ (\blue{3.51}) & $64.04_{\pm 0.03}$ (\blue{4.19}) & $15.13_{\pm 1.47}$ (\blue{1.73}) & $4.73$ & $36.38_{\pm 0.87}$ (\blue{1.42}) & $95.22_{\pm 0.26}$ (\blue{4.58}) & $52.99_{\pm 1.45}$ (\blue{2.75}) & $15.57_{\pm 0.47}$ (\blue{0.78}) & $5.58$ \\
        \bottomrule
        \end{tabular}
    }
\end{table}

We present a comprehensive comparison between our proposed method (LTU) and 9 other top-performing representative baselines in random sample unlearning, and results are shown in \cref{tab:main}. 
We follow the evaluation setting of \cite{fan2023salun} and tested on two forgetting ratios. 
With $10\%$ forgetting ratio, we can simulate the normally assumed scenario where data to be unlearned is a tiny subset of the whole dataset, while with $50\%$ forgetting ratio, we can simulate the scenario where unlearning algorithms are applied to forget a significant portion of data. 

It is worth noting that in \cref{tab:main} we ensure the baselines we compare with have full access to the remaining set, while our method only has access to 30\% of the remaining set (i.e., $\rho=0.3$).
Despite this, our proposed method still achieves state-of-the-art results consistently across all metrics, while maintaining a reasonably small unlearning time (UT). 
Specifically, the top-performing training-free approaches (\textit{Fisher} and \textit{SSD}) tend to severely damage the model utility when dealing with larger forgetting ratios.
On the other hand, as compared to the training or distillation-based methods, we consistently achieve better unlearning accuracy while using much less data from the remaining set, which shows the efficacy of our LTU framework.

\subsection{Ablation Studies}

To further investigate the efficacy of our proposed LTU framework, we perform ablation studies on CIFAR10 with ResNet-18 in random sample unlearning.

\begin{table}[t]
    \caption{
    Impact of partial access to the remaining set data at unlearning time.\\
    $\rho$ represents the percentage of data in $\mathcal{D}_r$ that is accessed during unlearning.
    }
    \label{tab:partial}
    \centering
    \resizebox{\textwidth}{!}{
        \begin{tabular}{c|ccccc|ccccc}
        \midrule
        \multirow{2}{*}{\textbf{Methods}} & \multicolumn{5}{c|}{ \textbf{Partial Access to $\mathcal{D}_r$ with $\bm{\rho=0.1}$} (excl. Retain) }  & \multicolumn{5}{c}{ \textbf{Partial Access to $\mathcal{D}_r$ with $\bm{\rho=0.3}$}  (excl. Retain) } \\
        & \multicolumn{1}{c|}{UA (\blue{$\Delta \downarrow$})} & \multicolumn{1}{c|}{RA (\blue{$\Delta \downarrow$})} & \multicolumn{1}{c|}{TA (\blue{$\Delta \downarrow$})} & \multicolumn{1}{c|}{MI (\blue{$\Delta \downarrow$})} & UT $\downarrow$ & \multicolumn{1}{c|}{UA (\blue{$\Delta \downarrow$})} & \multicolumn{1}{c|}{RA (\blue{$\Delta \downarrow$})} & \multicolumn{1}{c|}{TA (\blue{$\Delta \downarrow$})} & \multicolumn{1}{c|}{MI (\blue{$\Delta \downarrow$})} & UT $\downarrow$ \\
        \midrule
        Retrain ($\rho=1$) & $5.24_{\pm 0.69}$ (\blue{0.00}) & $100.00_{\pm 0.00}$ (\blue{0.00}) & $94.06_{\pm 0.02}$ (\blue{0.00}) & $12.88_{\pm 0.09}$ (\blue{0.00}) & $43.29$ & $5.24_{\pm 0.69}$ (\blue{0.00}) & $100.00_{\pm 0.00}$ (\blue{0.00}) & $94.06_{\pm 0.02}$ (\blue{0.00}) & $12.88_{\pm 0.09}$ (\blue{0.00}) & $43.29$ \\
        \midrule
        FT & $10.61_{\pm 0.29}$ (\blue{5.37}) & $88.01_{\pm 0.01}$ (\blue{11.99}) & $85.57_{\pm 0.88}$ (\blue{8.49}) & $9.04_{\pm 0.55}$ (\blue{3.84}) & $1.09$ & $9.65_{\pm 0.06}$ (\blue{4.41}) & $88.97_{\pm 0.85}$ (\blue{11.03}) & $87.99_{\pm 0.74}$ (\blue{6.07}) & $9.23_{\pm 0.38}$ (\blue{3.65}) & $1.42$ \\
        RandL & $20.42_{\pm 0.78}$ (\blue{15.18}) & $80.55_{\pm 0.96}$ (\blue{19.45}) & $78.91_{\pm 0.21}$ (\blue{15.15}) & $38.40_{\pm 0.21}$ (\blue{25.52}) & $1.15$ & $13.29_{\pm 0.59}$ (\blue{8.05}) & $87.16_{\pm 0.65}$ (\blue{12.84}) & $80.62_{\pm 0.66}$ (\blue{13.44}) & $35.16_{\pm 0.46}$ (\blue{22.28}) & $1.83$ \\
        BadT~\cite{chundawat2023badt} & $59.93_{\pm 1.07}$ (\blue{54.69}) & $39.49_{\pm 0.38}$ (\blue{60.51}) & $40.04_{\pm 0.14}$ (\blue{54.02}) & $25.50_{\pm 0.51}$ (\blue{12.62}) & $1.08$ & $37.17_{\pm 0.14}$ (\blue{31.93}) & $62.97_{\pm 0.04}$ (\blue{37.03}) & $59.02_{\pm 0.96}$ (\blue{35.04}) & $23.31_{\pm 0.13}$ (\blue{10.43}) & $1.23$ \\
        SalUn~\cite{fan2023salun} & $4.09_{\pm 0.47}$ (\blue{1.15}) & $90.97_{\pm 0.30}$ (\blue{9.03}) & $86.81_{\pm 0.80}$ (\blue{7.25}) & $14.72_{\pm 0.38}$ (\blue{1.84}) & $1.32$ & $4.88_{\pm 0.06}$ (\blue{0.36}) & $92.28_{\pm 0.04}$ (\blue{7.72}) & $91.11_{\pm 0.52}$ (\blue{2.95}) & $14.69_{\pm 0.09}$ (\blue{1.81}) & $2.48$ \\
        \midrule
        \textbf{LTU (Ours)} & $5.43_{\pm 0.41}$ (\blue{0.19}) & $92.33_{\pm 0.38}$ (\blue{7.67}) & $89.89_{\pm 0.10}$ (\blue{4.17}) & $12.34_{\pm 0.58}$ (\blue{0.54}) & $1.22$ & $4.37_{\pm 0.55}$ (\blue{0.87}) & $99.83_{\pm 0.09}$ (\blue{0.17}) & $93.95_{\pm 0.39}$ (\blue{0.31}) & $12.97_{\pm 0.02}$ (\blue{0.09}) & $2.63$ \\
        \bottomrule
        \end{tabular}
    }
\end{table}

\textbf{Impact of the size of partial remaining set}
Since it is often unrealistic to assume the remaining set ($\mathcal{D}_r$) is fully available at the unlearning time, our method aims to only utilize a small subset denoted as $\hat{\mathcal{D}_r}$ for unlearning in a post-hoc fashion (i.e. we do not require any prior setup nor saved information from the original model training process).
Here, we take a closer look at the impact of varying the size ($\rho$) of the partial remaining set for our method, as well as on the baselines that use the remaining set at the unlearning time (FT, RandL, BadT, SalUn). Note that we excluded baselines that rely on original training information or the full remaining set (Amnesiac, Fisher, SSD).
We report the results in \cref{tab:partial}.
We observe that our LTU framework consistently achieves significantly better overall unlearning performance when only a partial remaining set is provided.
Furthermore, our improvement over the baselines is much more significant when little data is provided (i.e., $\rho = 0.1$), as compared to when moderate amounts of data are provided (i.e., $\rho = 0.3$), thus showing our framework's efficacy at improving generalization with less data.

\textbf{Impact of unlearning feedback through meta optimization}
In our LTU framework, we apply \textit{meta-optimization scheme} twice and obtain two generalized feedback w.r.t. improving ``\textit{forgetting}'' and ``\textit{remembering}'' respectively. 
To evaluate the efficacy of this optimized feedback through meta-optimization in LTU, we test three variants of our framework: 
\circled{1} \textbf{LTU w/o ForFeed}: we remove the ``\textit{forgetting}'' feedback;
\circled{2} \textbf{LTU w/o RemFeed}: we remove the ``\textit{remembering}'' feedback;
\circled{3} \textbf{LTU w/o MetaOpt}: we use conventional training feedback only instead of our meta optimization scheme, i.e., meta-tune, test, and update.
As shown in \cref{tab:feedback}, these variants show a large performance gap when compared with our LTU framework, showing the efficacy of meta optimization.

\begin{figure}[!t] 
\begin{minipage}[t]{\textwidth}
  \begin{minipage}[t]{0.48\textwidth}
    \footnotesize
    \centering
    \makeatletter\def\@captype{table}\makeatother\caption{Impact of our meta optimization scheme.
    }
    \resizebox{\textwidth}{!}{%
        \begin{tabular}{c|ccccc}
        \midrule
        \multirow{2}{*}{\textbf{Methods}} & \multicolumn{5}{c}{\textbf{ ResNet-18 (CIFAR10) Random Sample Unlearning (10\%) }}  \\
        & \multicolumn{1}{c}{UA (\blue{$\Delta \downarrow$})} & \multicolumn{1}{c}{RA (\blue{$\Delta \downarrow$})} & \multicolumn{1}{c}{TA (\blue{$\Delta \downarrow$})} & \multicolumn{1}{c}{MI (\blue{$\Delta \downarrow$})} & UT $\downarrow$ \\
        \midrule
        Retrain & $5.24_{\pm 0.69}$ (\blue{0.00}) & $100.00_{\pm 0.00}$ (\blue{0.00}) & $94.26_{\pm 0.02}$ (\blue{0.00}) & $12.88_{\pm 0.09}$ (\blue{0.00}) & $43.29$ \\
        \midrule
        LTU (w/o ForFeed) & $11.97_{\pm 0.21}$ (\blue{6.73}) & $98.34_{\pm 0.59}$ (\blue{1.66}) & $93.21_{\pm 0.38}$ (\blue{1.05}) & $28.81_{\pm 0.37}$ (\blue{15.93}) & $1.53$ \\
        LTU (w/o RemFeed) & $23.35_{\pm 3.83}$ (\blue{18.11}) & $72.30_{\pm 3.44}$ (\blue{27.70}) & $70.51_{\pm 4.24}$ (\blue{23.75}) & $25.16_{\pm 0.12}$ (\blue{12.28}) & $0.86$ \\
        LTU (w/o MetaOpt) & $18.09_{\pm 1.62}$ (\blue{12.85}) & $91.24_{\pm 1.13}$ (\blue{8.76}) & $90.22_{\pm 1.13}$ (\blue{4.04}) & $16.56_{\pm 0.19}$ (\blue{3.68}) & $2.41$ \\
        LTU (Ours) & $4.37_{\pm 0.55}$ (\blue{0.87}) & $99.83_{\pm 0.09}$ (\blue{0.17}) & $93.95_{\pm 0.39}$ (\blue{0.31}) & $12.97_{\pm 0.02}$ (\blue{0.09}) & $2.63$ \\
        \bottomrule
    \end{tabular}
    }
    \label{tab:feedback}
\end{minipage}
\hspace{0.01\textwidth}
\begin{minipage}[t]{0.49\textwidth}
    \footnotesize
    \centering
    \makeatletter\def\@captype{table}\makeatother\caption{Impact of our Gradient Harmonization strategy.
    }  
    \resizebox{\textwidth}{!}{%
        \begin{tabular}{c|ccccc}
        \midrule
        \multirow{2}{*}{\textbf{Methods}} & \multicolumn{5}{c}{\textbf{ ResNet-18 (CIFAR10) Random Sample Unlearning (10\%) }}  \\
        & \multicolumn{1}{c}{UA (\blue{$\Delta \downarrow$})} & \multicolumn{1}{c}{RA (\blue{$\Delta \downarrow$})} & \multicolumn{1}{c}{TA (\blue{$\Delta \downarrow$})} & \multicolumn{1}{c}{MI (\blue{$\Delta \downarrow$})} & UT $\downarrow$ \\
        \midrule
        Retrain & $5.24_{\pm 0.69}$ (\blue{0.00}) & $100.00_{\pm 0.00}$ (\blue{0.00}) & $94.26_{\pm 0.02}$ (\blue{0.00}) & $12.88_{\pm 0.09}$ (\blue{0.00}) & $43.29$ \\
        \midrule
        LTU (w/ GradAdd) & $6.26_{\pm 0.55}$ (\blue{1.02}) & $97.40_{\pm 0.70}$ (\blue{2.60}) & $89.54_{\pm 0.98}$ (\blue{4.72}) & $15.80_{\pm 0.80}$ (\blue{2.92}) & $2.61$ \\
        LTU (w/ iterative Grad) & $7.11_{\pm 0.23}$ (\blue{1.87}) & $96.13_{\pm 0.04}$ (\blue{3.87}) & $87.51_{\pm 0.28}$ (\blue{6.75}) & $16.03_{\pm 0.21}$ (\blue{3.15}) & $2.77$ \\
        \textbf{LTU (Ours)} & $4.37_{\pm 0.55}$ (\blue{0.87}) & $99.83_{\pm 0.09}$ (\blue{0.17}) & $93.95_{\pm 0.39}$ (\blue{0.31}) & $12.97_{\pm 0.02}$ (\blue{0.09}) & $2.63$ \\
        \bottomrule
    \end{tabular}
    }
    \label{tab:harmonization}
\end{minipage}

\end{minipage}
\end{figure}

\textbf{Impact of Gradient Harmonization}
In our LTU framework, we introduce a Gradient Harmonization strategy to harmonize the two optimized gradients obtained through meta-optimization w.r.t. ``\textit{forgetting}'' and ``\textit{remembering}''. 
To evaluate the effectiveness of this strategy, we implement two variants of LTU: 
\circled{1} \textbf{LTU w/ GradAdd}: We replace Gradient Harmonization with naive gradient addition between the ``\textit{remembering}'' and ``\textit{forgetting}'' gradients.
\circled{2} \textbf{LTU w/ iterative Grad}: We remove the Gradient Harmonization operation and instead of obtaining both gradients every iteration, we obtain ``\textit{forgetting}'' feedback at odd iterations and ``\textit{remembering}'' feedback at even iterations alternatively.
We report the results in \cref{tab:harmonization}, where our method significantly outperforms the baselines, which verifies the efficacy of our Gradient Harmonization strategy.


\section{Conclusion}
In this paper, we propose LTU framework from a novel perspective of meta learning. 
LTU simultaneously obtains \textit{generalizable} feedback towards \textit{forgetting} and \textit{remembering} in a unified manner, while using only a small amount of remaining data. Overall our LTU framework consistently achieves state-of-the-art results, showing its efficacy and practicality in real-world applications.

\ \\
\textbf{Acknowledgments} 
Special thanks to Duo Peng and Li Xu for their valuable advice and input. This research is partly supported by the Ministry of Education, Singapore, under the AcRF Tier 2 Projects (MOE-T2EP20222-0009 and MOE-T2EP20123-0014), as well as the National Research Foundation Singapore through its AI Singapore Programme (AISG-100E-2023-121).

\bibliographystyle{splncs04}
\bibliography{egbib}

\begin{thebibliography}{10}
\providecommand{\url}[1]{\texttt{#1}}
\providecommand{\urlprefix}{URL }
\providecommand{\doi}[1]{https://doi.org/#1}

\bibitem{antoniou2019learning}
Antoniou, A., Storkey, A.J.: Learning to learn by self-critique. Advances in Neural Information Processing Systems  \textbf{32} (2019)

\bibitem{blanco2024digital}
Blanco-Justicia, A., Jebreel, N., Manzanares, B., S{\'a}nchez, D., Domingo-Ferrer, J., Collell, G., Tan, K.E.: Digital forgetting in large language models: A survey of unlearning methods. arXiv preprint arXiv:2404.02062  (2024)

\bibitem{SISA}
Bourtoule, L., Chandrasekaran, V., Choquette-Choo, C.A., Jia, H., Travers, A., Zhang, B., Lie, D., Papernot, N.: Machine unlearning. In: 2021 IEEE Symposium on Security and Privacy (SP). pp. 141--159. IEEE (2021)

\bibitem{cha2024learning}
Cha, S., Cho, S., Hwang, D., Lee, H., Moon, T., Lee, M.: Learning to unlearn: Instance-wise unlearning for pre-trained classifiers. In: Proceedings of the AAAI Conference on Artificial Intelligence. vol.~38, pp. 11186--11194 (2024)

\bibitem{chen2023boundary}
Chen, M., Gao, W., Liu, G., Peng, K., Wang, C.: Boundary unlearning: Rapid forgetting of deep networks via shifting the decision boundary. In: Proceedings of the IEEE/CVF Conference on Computer Vision and Pattern Recognition. pp. 7766--7775 (2023)

\bibitem{chen2021machine}
Chen, M., Zhang, Z., Wang, T., Backes, M., Humbert, M., Zhang, Y.: When machine unlearning jeopardizes privacy. In: Proceedings of the 2021 ACM SIGSAC conference on computer and communications security. pp. 896--911 (2021)

\bibitem{chen2024fast}
Chen, R., Yang, J., Xiong, H., Bai, J., Hu, T., Hao, J., Feng, Y., Zhou, J.T., Wu, J., Liu, Z.: Fast model debias with machine unlearning. Advances in Neural Information Processing Systems  \textbf{36} (2024)

\bibitem{cheng2023gnndelete}
Cheng, J., Dasoulas, G., He, H., Agarwal, C., Zitnik, M.: Gnndelete: A general strategy for unlearning in graph neural networks. arXiv preprint arXiv:2302.13406  (2023)

\bibitem{chourasia2022forget}
Chourasia, R., Shah, N., Shokri, R.: Forget unlearning: Towards true data-deletion in machine learning. arXiv preprint arXiv:2210.08911  (2022)

\bibitem{chundawat2023badt}
Chundawat, V.S., Tarun, A.K., Mandal, M., Kankanhalli, M.: Can bad teaching induce forgetting? unlearning in deep networks using an incompetent teacher. In: Proceedings of the AAAI Conference on Artificial Intelligence. vol.~37, pp. 7210--7217 (2023)

\bibitem{deng2009imagenet}
Deng, J., Dong, W., Socher, R., Li, L.J., Li, K., Fei-Fei, L.: Imagenet: A large-scale hierarchical image database. In: 2009 IEEE conference on computer vision and pattern recognition. pp. 248--255. Ieee (2009)

\bibitem{di2024adversarial}
Di, Z., Yu, S., Vorobeychik, Y., Liu, Y.: Adversarial machine unlearning. arXiv preprint arXiv:2406.07687  (2024)

\bibitem{dosovitskiy2020image}
Dosovitskiy, A., Beyer, L., Kolesnikov, A., Weissenborn, D., Zhai, X., Unterthiner, T., Dehghani, M., Minderer, M., Heigold, G., Gelly, S., et~al.: An image is worth 16x16 words: Transformers for image recognition at scale. arXiv preprint arXiv:2010.11929  (2020)

\bibitem{dukler2023safe}
Dukler, Y., Bowman, B., Achille, A., Golatkar, A., Swaminathan, A., Soatto, S.: Safe: Machine unlearning with shard graphs. arXiv preprint arXiv:2304.13169  (2023)

\bibitem{fan2023salun}
Fan, C., Liu, J., Zhang, Y., Wei, D., Wong, E., Liu, S.: Salun: Empowering machine unlearning via gradient-based weight saliency in both image classification and generation. arXiv preprint arXiv:2310.12508  (2023)

\bibitem{MAML}
Finn, C., Abbeel, P., Levine, S.: Model-agnostic meta-learning for fast adaptation of deep networks. In: International conference on machine learning. pp. 1126--1135. PMLR (2017)

\bibitem{foo2023system}
Foo, L.G., Gong, J., Fan, Z., Liu, J.: System-status-aware adaptive network for online streaming video understanding. In: Proceedings of the IEEE/CVF Conference on Computer Vision and Pattern Recognition. pp. 10514--10523 (2023)

\bibitem{foo2022era}
Foo, L.G., Li, T., Rahmani, H., Ke, Q., Liu, J.: Era: Expert retrieval and assembly for early action prediction. In: European Conference on Computer Vision. pp. 670--688. Springer (2022)

\bibitem{foster2023ssd}
Foster, J., Schoepf, S., Brintrup, A.: Fast machine unlearning without retraining through selective synaptic dampening. arXiv preprint arXiv:2308.07707  (2023)

\bibitem{golatkar2021mixed}
Golatkar, A., Achille, A., Ravichandran, A., Polito, M., Soatto, S.: Mixed-privacy forgetting in deep networks. In: Proceedings of the IEEE/CVF conference on computer vision and pattern recognition. pp. 792--801 (2021)

\bibitem{golatkar2020eternal}
Golatkar, A., Achille, A., Soatto, S.: Eternal sunshine of the spotless net: Selective forgetting in deep networks. In: Proceedings of the IEEE/CVF Conference on Computer Vision and Pattern Recognition. pp. 9304--9312 (2020)

\bibitem{golatkar2020forgetting}
Golatkar, A., Achille, A., Soatto, S.: Forgetting outside the box: Scrubbing deep networks of information accessible from input-output observations. In: Computer Vision--ECCV 2020: 16th European Conference, Glasgow, UK, August 23--28, 2020, Proceedings, Part XXIX 16. pp. 383--398. Springer (2020)

\bibitem{graves2021amnesiac}
Graves, L., Nagisetty, V., Ganesh, V.: Amnesiac machine learning. In: Proceedings of the AAAI Conference on Artificial Intelligence. vol.~35, pp. 11516--11524 (2021)

\bibitem{guo2019certified}
Guo, C., Goldstein, T., Hannun, A., Van Der~Maaten, L.: Certified data removal from machine learning models. arXiv preprint arXiv:1911.03030  (2019)

\bibitem{guo2020learning}
Guo, J., Zhu, X., Zhao, C., Cao, D., Lei, Z., Li, S.Z.: Learning meta face recognition in unseen domains. In: Proceedings of the IEEE/CVF Conference on Computer Vision and Pattern Recognition. pp. 6163--6172 (2020)

\bibitem{resnet}
He, K., Zhang, X., Ren, S., Sun, J.: Deep residual learning for image recognition. In: Proceedings of the IEEE conference on computer vision and pattern recognition. pp. 770--778 (2016)

\bibitem{heng2024selective}
Heng, A., Soh, H.: Selective amnesia: A continual learning approach to forgetting in deep generative models. Advances in Neural Information Processing Systems  \textbf{36} (2024)

\bibitem{hu2022survey}
Hu, H., Salcic, Z., Sun, L., Dobbie, G., Yu, P.S., Zhang, X.: Membership inference attacks on machine learning: A survey. ACM Computing Surveys (CSUR)  \textbf{54}(11s),  1--37 (2022)

\bibitem{huang2021metasets}
Huang, C., Cao, Z., Wang, Y., Wang, J., Long, M.: Metasets: Meta-learning on point sets for generalizable representations. In: Proceedings of the IEEE/CVF Conference on Computer Vision and Pattern Recognition. pp. 8863--8872 (2021)

\bibitem{kim2022efficient}
Kim, J., Woo, S.S.: Efficient two-stage model retraining for machine unlearning. In: Proceedings of the IEEE/CVF Conference on Computer Vision and Pattern Recognition. pp. 4361--4369 (2022)

\bibitem{koch2023no}
Koch, K., Soll, M.: No matter how you slice it: Machine unlearning with sisa comes at the expense of minority classes. In: 2023 IEEE Conference on Secure and Trustworthy Machine Learning (SaTML). pp. 622--637. IEEE (2023)

\bibitem{kodge2023deep}
Kodge, S., Saha, G., Roy, K.: Deep unlearning: Fast and efficient training-free approach to controlled forgetting. arXiv preprint arXiv:2312.00761  (2023)

\bibitem{koh2023disposable}
Koh, S., Shon, H., Lee, J., Hong, H.G., Kim, J.: Disposable transfer learning for selective source task unlearning. In: Proceedings of the IEEE/CVF International Conference on Computer Vision. pp. 11752--11760 (2023)

\bibitem{CIFAR100}
Krizhevsky, A., Hinton, G., et~al.: Learning multiple layers of features from tiny images  (2009)

\bibitem{kurmanji2024towards}
Kurmanji, M., Triantafillou, P., Hayes, J., Triantafillou, E.: Towards unbounded machine unlearning. Advances in neural information processing systems  \textbf{36} (2024)

\bibitem{li2024machine}
Li, G., Hsu, H., Marculescu, R., et~al.: Machine unlearning for image-to-image generative models. arXiv preprint arXiv:2402.00351  (2024)

\bibitem{lin2023erm}
Lin, S., Zhang, X., Chen, C., Chen, X., Susilo, W.: Erm-ktp: Knowledge-level machine unlearning via knowledge transfer. In: Proceedings of the IEEE/CVF Conference on Computer Vision and Pattern Recognition. pp. 20147--20155 (2023)

\bibitem{liu2023muter}
Liu, J., Xue, M., Lou, J., Zhang, X., Xiong, L., Qin, Z.: Muter: Machine unlearning on adversarially trained models. In: Proceedings of the IEEE/CVF International Conference on Computer Vision. pp. 4892--4902 (2023)

\bibitem{liu2022membership}
Liu, Y., Zhao, Z., Backes, M., Zhang, Y.: Membership inference attacks by exploiting loss trajectory. In: Proceedings of the 2022 ACM SIGSAC Conference on Computer and Communications Security. pp. 2085--2098 (2022)

\bibitem{mantelero2013eu}
Mantelero, A.: The eu proposal for a general data protection regulation and the roots of the ‘right to be forgotten’. Computer Law \& Security Review  \textbf{29}(3),  229--235 (2013)

\bibitem{mehta2022deep}
Mehta, R., Pal, S., Singh, V., Ravi, S.N.: Deep unlearning via randomized conditionally independent hessians. In: Proceedings of the IEEE/CVF Conference on Computer Vision and Pattern Recognition. pp. 10422--10431 (2022)

\bibitem{nguyen2020variational}
Nguyen, Q.P., Low, B.K.H., Jaillet, P.: Variational bayesian unlearning. Advances in Neural Information Processing Systems  \textbf{33},  16025--16036 (2020)

\bibitem{nichol2018first}
Nichol, A., Achiam, J., Schulman, J.: On first-order meta-learning algorithms. arXiv preprint arXiv:1803.02999  (2018)

\bibitem{pan2023gradmdm}
Pan, J., Foo, L.G., Zheng, Q., Fan, Z., Rahmani, H., Ke, Q., Liu, J.: Gradmdm: Adversarial attack on dynamic networks. IEEE Transactions on Pattern Analysis and Machine Intelligence  \textbf{45}(9),  11374--11381 (2023)

\bibitem{peng2024joint}
Peng, D., Xu, L., Ke, Q., Hu, P., Liu, J.: Joint attribute and model generalization learning for privacy-preserving action recognition. Advances in Neural Information Processing Systems  \textbf{36} (2024)

\bibitem{peng2024harnessing}
Peng, D., Zhang, Z., Hu, P., Ke, Q., Yau, D., Liu, J.: Harnessing text-to-image diffusion models for category-agnostic pose estimation. In: European Conference on Computer Vision. Springer (2024)

\bibitem{peste2021ssse}
Peste, A., Alistarh, D., Lampert, C.H.: Ssse: Efficiently erasing samples from trained machine learning models. arXiv preprint arXiv:2107.03860  (2021)

\bibitem{rajeswaran2019meta}
Rajeswaran, A., Finn, C., Kakade, S.M., Levine, S.: Meta-learning with implicit gradients. Advances in neural information processing systems  \textbf{32} (2019)

\bibitem{salem2018ml}
Salem, A., Zhang, Y., Humbert, M., Berrang, P., Fritz, M., Backes, M.: Ml-leaks: Model and data independent membership inference attacks and defenses on machine learning models. arXiv preprint arXiv:1806.01246  (2018)

\bibitem{shen2024label}
Shen, S., Zhang, C., Zhao, Y., Bialkowski, A., Chen, W., Xu, M.: Label-agnostic forgetting: A supervision-free unlearning in deep models. arXiv preprint arXiv:2404.00506  (2024)

\bibitem{shokri2017membership}
Shokri, R., Stronati, M., Song, C., Shmatikov, V.: Membership inference attacks against machine learning models. In: 2017 IEEE symposium on security and privacy (SP). pp. 3--18. IEEE (2017)

\bibitem{snell2017prototypical}
Snell, J., Swersky, K., Zemel, R.: Prototypical networks for few-shot learning. Advances in neural information processing systems  \textbf{30} (2017)

\bibitem{tarun2023fast}
Tarun, A.K., Chundawat, V.S., Mandal, M., Kankanhalli, M.: Fast yet effective machine unlearning. IEEE Transactions on Neural Networks and Learning Systems  (2023)

\bibitem{thudi2022unrolling}
Thudi, A., Deza, G., Chandrasekaran, V., Papernot, N.: Unrolling sgd: Understanding factors influencing machine unlearning. In: 2022 IEEE 7th European Symposium on Security and Privacy (EuroS\&P). pp. 303--319. IEEE (2022)

\bibitem{wu2020deltagrad}
Wu, Y., Dobriban, E., Davidson, S.: Deltagrad: Rapid retraining of machine learning models. In: International Conference on Machine Learning. pp. 10355--10366. PMLR (2020)

\bibitem{xu2023machine}
Xu, H., Zhu, T., Zhang, L., Zhou, W., Yu, P.S.: Machine unlearning: A survey. ACM Comput. Surv.  \textbf{56}(1) (2023)

\bibitem{xu2024machine}
Xu, J., Wu, Z., Wang, C., Jia, X.: Machine unlearning: Solutions and challenges. IEEE Transactions on Emerging Topics in Computational Intelligence  (2024)

\bibitem{MCRES_2023_CVPR}
Xu, L., Huang, M.H., Shang, X., Yuan, Z., Sun, Y., Liu, J.: Meta compositional referring expression segmentation. In: Proceedings of the IEEE/CVF Conference on Computer Vision and Pattern Recognition. pp. 19478--19487 (2023)

\bibitem{xu2023experts}
Xu, L., Liu, J.: Experts collaboration learning for continual multi-modal reasoning. IEEE Transactions on Image Processing  (2023)

\bibitem{ye2022learning}
Ye, J., Fu, Y., Song, J., Yang, X., Liu, S., Jin, X., Song, M., Wang, X.: Learning with recoverable forgetting. In: European Conference on Computer Vision. pp. 87--103. Springer (2022)

\bibitem{yu2020gradient}
Yu, T., Kumar, S., Gupta, A., Levine, S., Hausman, K., Finn, C.: Gradient surgery for multi-task learning. arXiv preprint arXiv:2001.06782  (2020)

\bibitem{zhang2024geniu}
Zhang, C., Shen, S., Zhao, Y., Chen, W.T., Xu, M.: Geniu: A restricted data access unlearning for imbalanced data. arXiv preprint arXiv:2406.07885  (2024)

\bibitem{zhang2023review}
Zhang, H., Nakamura, T., Isohara, T., Sakurai, K.: A review on machine unlearning. SN Computer Science  \textbf{4}(4), ~337 (2023)

\bibitem{zhang2022distilling}
Zhang, Z., Zhou, C., Tu, Z.: Distilling inter-class distance for semantic segmentation. arXiv preprint arXiv:2205.03650  (2022)

\bibitem{zhang2022prompt}
Zhang, Z., Zhou, Y., Zhao, X., Che, T., Lyu, L.: Prompt certified machine unlearning with randomized gradient smoothing and quantization. Advances in Neural Information Processing Systems  \textbf{35},  13433--13455 (2022)

\bibitem{zhou2023unified}
Zhou, J., Li, H., Liao, X., Zhang, B., He, W., Li, Z., Zhou, L., Gao, X.: A unified method to revoke the private data of patients in intelligent healthcare with audit to forget. Nature Communications  \textbf{14}(1), ~6255 (2023)

\end{thebibliography}
\end{document}